\documentclass[runningheads]{llncs}
\usepackage[T1]{fontenc}
\usepackage{graphicx,verbatim}
\usepackage{booktabs}
\usepackage{longtable}
\usepackage{marvosym}
\usepackage{hyperref}

\newcommand{\method}{BioVFM}

\begin{document}
\title{BioVFM-21M: Benchmarking and Scaling Self-Supervised Vision Foundation Models \\for Biomedical Image Analysis}

\author{
    Jiarun Liu\inst{1,2,3} \and
    Hong-Yu Zhou \and
    Weijian Huang\inst{1,2,3} \and
    Hao Yang\inst{1,2,3} \and \\
    Dongning Song\inst{1,2,3} \and
    Tao Tan\inst{4} \and
    Yong Liang\inst{3} \and
    Shanshan Wang\inst{1}\textsuperscript{(\Letter)}
}

\authorrunning{Liu. Jiarun et al.}
%
\institute{Paul C. Lauterbur Research Center for Biomedical Imaging, Shenzhen Institutes of \\
Advanced Technology, Chinese Academy of Sciences, Shenzhen, China\\
\email{ss.wang@siat.ac.cn}\and
Pengcheng Laboratory, Shenzhen, China \and
University of Chinese Academy of Sciences, Beijing, China \and
Faculty of Applied Sciences, Macao Polytechnic University, Macao, China
}

\maketitle              

\begin{abstract}
Scaling up model and data size have demonstrated impressive performance improvement over a wide range of tasks. Despite extensive studies on scaling behaviors for general-purpose tasks, medical images exhibit substantial differences from natural data. It remains unclear the key factors in developing medical vision foundation models at scale due to the absence of an extensive understanding of scaling behavior in the medical domain. In this paper, we explored the scaling behavior across model sizes, training algorithms, data sizes, and imaging modalities in developing scalable medical vision foundation models by self-supervised learning. To support scalable pretraining, we introduce \method{}-21M, a large-scale biomedical image dataset encompassing a wide range of biomedical image modalities and anatomies. We observed that scaling up does provide benefits but varies across tasks. Additional analysis reveals several factors correlated with scaling benefits. Finally, we propose \method{}, a large-scale medical vision foundation model pretrained on 21 million biomedical images, which outperforms the previous state-of-the-art foundation models across 12 medical benchmarks. Our results highlight that while scaling up is beneficial for pursuing better performance, task characteristics, data diversity, pretraining methods, and computational efficiency remain critical considerations for developing scalable medical foundation models.

\end{abstract}

\section{Introduction}
Biomedical imaging is at the forefront of healthcare, playing an essential role in the diagnosis and treatment of diseases \cite{azad_foundational_2023,wang2021annotation}. Foundation models (FMs) offer a more efficient alternative that allows straightforward application to a wide range of tasks with minimal labeled data \cite{zhang_challenges_2024,azad_foundational_2023}. The success of FMs is largely driven by the scalable self-supervised approaches \cite{dinov2,mae}, which enable generalizable representations from large datasets without annotations \cite{yang2024afloc}. The well-known scaling law \cite{kaplan2020scaling,hoffmann2022scaling} implies that the model performance can be continuously improved with increased model size, data size, and amount of computes. However, training large foundation models is expensive \cite{hagele2024scaling} -- in computation, energy, time, and data. Unfortunately, most of the existing scaling studies are drawn from the general domain \cite{kaplan2020scaling,hoffmann2022scaling,hagele2024scaling,alabdulmohsin2023vitshape,zhai2022scaling_vit}, while biomedical data exhibit distinct characteristics compared to general data \cite{liu2024mamba,zhou2023unified}. Considering the diminishing returns and substantial resource requirements \cite{hoffmann2022scaling}, it is important to understand the scalability in the medical domain before further scaling up.

Nevertheless, large-scale scalability research in biomedicine requires sufficient training data \cite{kaplan2020scaling}. Despite the superior performance, many existing medical FMs and datasets are developed with a focus on a specific modality \cite{raddino,zhou2022generalized,zhou2023transformer,huang2024enhancing,huang2024mia,vorontsov2024foundation}, organ \cite{zhou2023foundation,liu2024brain}, or task \cite{ma2024segment,butoi2023universeg}. Such a focus, by its nature, maintains the existing dataset biases \cite{jones2024causal} that are limited to specific biomedical domains and reduces the opportunities to combine more data from diverse modalities \cite{moor2023foundation}, thus limiting the diversity and quantity of training data. The model may struggle to learn generalizable representations across different modalities and anatomical structures \cite{moor2023foundation}, which limits the applicability of learned scaling behavior to a wider spectrum of medical applications. 

In this work, we explore the scaling behavior for developing self-supervised medical vision FMs. Our empirical experiments reveal that scaling up did take benefits but differed across tasks, whereby optimal scaling strategy depends on task and dataset characteristics. Our contributions are summarized as follows:
\begin{itemize}
    \item We curate \method{}-21M, a large biomedical image dataset with over 21 million images, spanning 10 imaging modalities and 30 anatomical structures.
    \item We conducted extensive experiments with an additional correlation analysis in exploring the scalability across model sizes, training algorithms, data sizes, and imaging modalities in developing self-supervised medical vision FMs. 
    \item We propose \method{}, a self-supervised medical vision foundation model at scale, demonstrating remarkable performance across 12 medical benchmarks.
\end{itemize}

\begin{figure}[t]
    \centering
    \includegraphics[width=\linewidth]{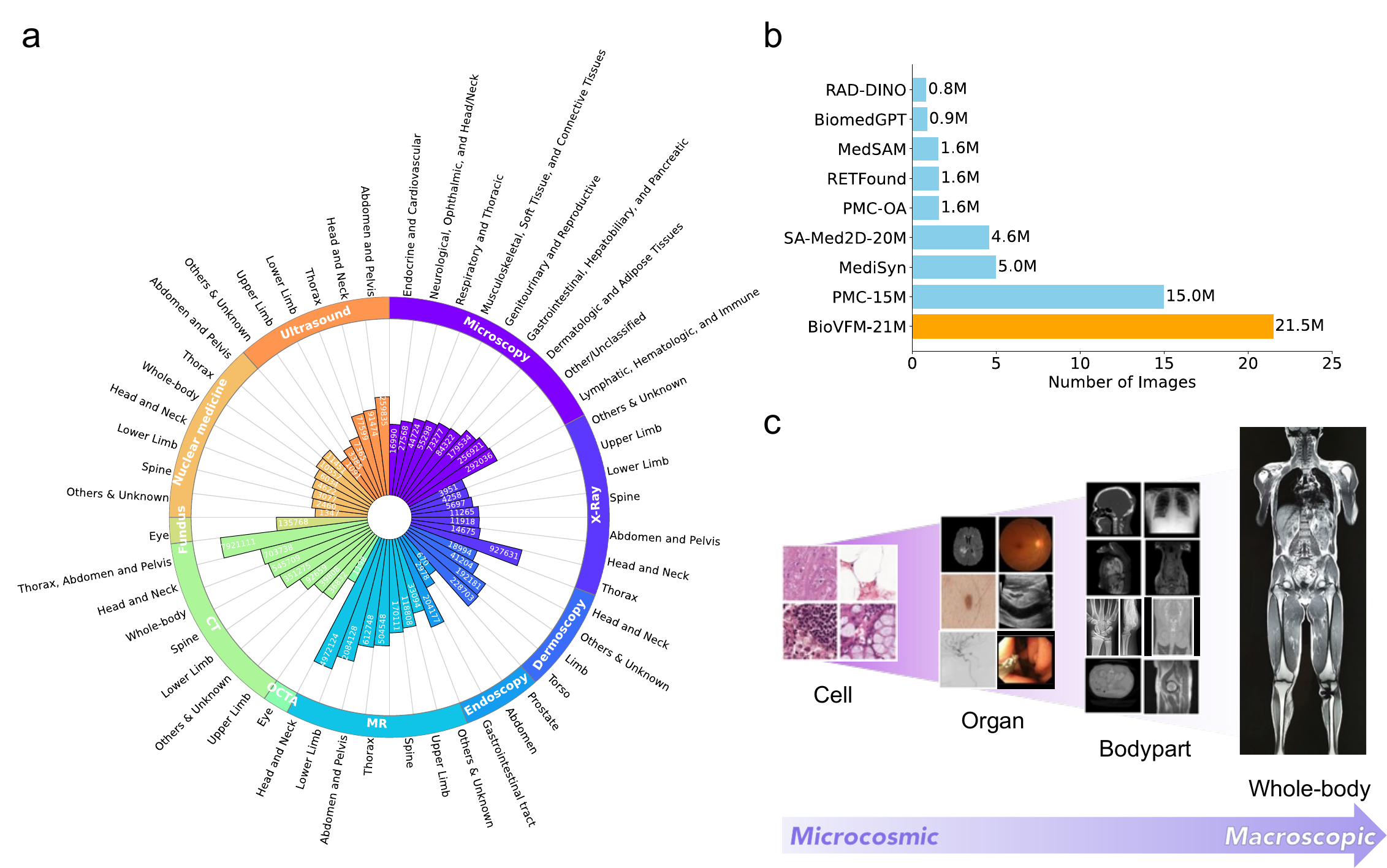}
    \caption{
        \textbf{The proposed \method{}-21M dataset.} 
        \textbf{a.} \method{}-21M covers a wide range of imaging modalities and anatomical areas. The illustration is exponential-scaled.
        \textbf{b.} Comparison with existing large-scale biomedical image datasets. 
        \textbf{c.} \method{}-21M covers images from microcosmic to macroscopic. 
    }
    \label{fig:dataset_demo}
\end{figure}

\section{The \method{}-21M dataset}
To support the training of \method{}, we curated \method{}-21M, a large biomedical image dataset consisting of 21 million biomedical images from over 43 publicly available data sources. \method{}-21M following three key principles: 
\textit{1) large-scale}: As shown in Fig. \ref{fig:dataset_demo}b, \method{}-21M is composed of 21 million images, which is significantly larger than previous datasets sourced from medical image datasets or medical websites.
\textit{2) data diversity}: As illustrated in Fig. \ref{fig:dataset_demo}a and \ref{fig:dataset_demo}c, \method{}-21M cover most of the major anatomical regions of the human body from microcosmic to macroscopic across 10 biomedical imaging modalities, including CT, MR, Microscopy, X-Ray, Dermoscopy, Ultrasound, Endoscopy, Fundus, Nuclear medicine, and OCTA. \textit{3) publicly available}: We focused on collecting data from publicly available sources to ensure transparency and accessibility. We will open the links of source datasets, processing codes, and more details of \method{}-21M in \url{https://github.com/JiarunLiu/BioVFM-21M}.

To reduce the risk of data leakage, we excluded 1.8 million testing and validation images based on default data splits of each source dataset. In addition, all data sources related to our evaluation benchmarks were excluded from \method{}-21M, ensuring that performance is not inflated by overlapped training and evaluation data. For 3D images, we slice them along axial, sagittal, and coronal views. The initial version of our dataset has over 30 million images, while all of the detected padding slices were removed since it does not have anatomical structure information. Furthermore, we removed images with aspect ratio smaller than 0.4. All images were normalized to 0-255 and resized to 224x224px while preserving their original aspect ratio.

\section{Method}
\subsection{Self-supervised pretraining}
We investigate the scalability of the self-supervised medical vision foundation models by varying model sizes, training algorithms, data numbers, and imaging modalities. Specifically, we select two representative self-supervised learning methods for model and algorithm scalability experiments: MAE \cite{mae} (\method{}-M) and DINO V2 \cite{dinov2} (\method{}-D). These methods are trained with model sizes ranging from 5 million to 303 million parameters while other settings follow the default settings as the author suggests. To optimize computational resources, we pretrain these models on a subset of 2.45 million images drawn from diverse imaging modalities and anatomical structures, sourced from \cite{pmc_clip,radimagenet}. 
In the data and modality scaling experiments, we extract two smaller subsets from \method{}-21M with different data numbers and image modalities, which are \method{}-US and \method{}-0.2M. The \method{}-US is an ultrasound subset with 0.4 million ultrasound images from \method{}-21M. Similarly, \method{}-0.2M is another small subset that was randomly sampled from \method{}-21M but contains all 10 image modalities from \method{}-21M. We train three models with 303 million parameters using \method{}-D on \method{}-US, \method{}-0.2M, and the full \method{}-21M dataset. All models are pretrained on a server with 8 NVIDIA A100 80GB GPUs.

\subsection{Evaluation benchmarks}
In contrast to previous scaling studies that evaluate loss values, we assess model performance using the Area Under the Curve (AUC) score across 12 diagnostic benchmarks from MedMNIST \cite{medmnist}. The 12 benchmarks span a variety of imaging modalities, anatomical regions, number of classes (ranging from 2 to 14), sample numbers (from 100 to 100,000), and task types (binary/multi-class classification, multi-label classification). For evaluation, we finetune the pretrained models using linear-probing settings with an NVIDIA 2080Ti GPU by default. We specify the learning rate by 0.01 with AdamW optimizer, a batch size of 128. The maximum epoch number is 100 and early stopping is applied based on validation loss. These hyperparameters were chosen based on preliminary experiments, which demonstrated optimal performance. In addition, we compare the performance of \method{} with ImageNet pretrained ViTs (IN21K-ViT-Base, IN21K-ViT-Large), RAD-DINO \cite{raddino}, BiomedCLIP \cite{biomedclip}, and BiomedGPT \cite{biomedclip}, which differ in pretraining data size (0.9 to 21 million), supervision (self-supervised learning (SSL), supervised learning (SL), vision-language pretraining (VLP)), model size (85.8 to 303 million), and data sources (natural or medical data).

\subsection{Scalability analysis}
To quantify the benefits of model scaling, we fit a power function $y=e^bx^a$ as proposed in \cite{porian2024resolving}. Here $a$ indicates the slope of the scaling curve and $b$ is the corresponding intersection. A higher slope $a$ indicates that scaling up the model size has a more positive effect on performance for the corresponding task. Additionally, to investigate the underlying factors related to scalability, we extract 4 quantitative metrics from downstream tasks: the number of samples, Davies-Bouldin Index (DBI) \cite{davies1979dbi}, the number of classes, and g-zip compressibility \cite{pandey2024gzip}. The number of samples and the number of classes provide fundamental insights into the dataset’s characteristics, while DBI reflects the complexity of data distribution and classification difficulty. The g-zip compressibility, on the other hand, quantifies the information redundancy of the data. With these metrics, we analyze the correlation between the computed values and the scaling slopes $a$ using the Pearson Correlation Coefficient (PCC) in Sec. \ref{sec:scaling_corr}.

\section{Results}
\subsection{BioVFM: Biomedical vision foundation model at scale}
Based on the \method{}-21M dataset, we propose \method{}, a Biomedical Vision Foundation Model at scale. By pretraining on 21 million biomedical images with 630 million parameters using \method{}-D, \method{} sets a new benchmark for generalizable medical foundation models. As shown in Table \ref{tab:final_mean_results}, \method{} significantly outperforms existing medical foundation model BiomedGPT \cite{biomedgpt}, BiomedCLIP \cite{biomedclip}, and RAD-DINO \cite{raddino} across 12 medical benchmarks with linear-probing by at least 3.32\% in MCC, 2.81\% in BA, 2.14\% in F1 score, and 0.94\% in AUC. Our results highlight the advantages of scalable self-supervised pretraining in the development of generalizable medical foundation models.

\begin{table}[t]
    \caption{\textbf{Average performance on 12 MedMNIST benchmarks.} The values in brackets indicate 95\% CI. $\dagger$: Developed with classification labels. $\ddagger$: Developed with image-text pairs. $*$: Developed for a specific medical domain. 
    MCC: Matthews Correlation Coefficient. BA: Balanced Accuracy.
    }
    \label{tab:final_mean_results}
    \centering
    \begin{tabular}{c|cccc}
        \toprule
        Model                        & MCC                                                                    & BA                                                                     & F1                                                                     & AUC                                                                    \\ \midrule
        IN21K-ViT-Base$^\dagger$               & \begin{tabular}[c]{@{}c@{}}61.59\\ (59.98,62.97)\end{tabular}          & \begin{tabular}[c]{@{}c@{}}69.46\\ (68.53,70.35)\end{tabular}          & \begin{tabular}[c]{@{}c@{}}67.16\\ (66.39,67.86)\end{tabular}          & \begin{tabular}[c]{@{}c@{}}91.86\\ (91.18,92.43)\end{tabular}          \\
        IN21K-ViT-Large$^\dagger$              & \begin{tabular}[c]{@{}c@{}}63.33\\ (61.78,64.73)\end{tabular}          & \begin{tabular}[c]{@{}c@{}}70.87\\ (69.83,71.80)\end{tabular}          & \begin{tabular}[c]{@{}c@{}}67.79\\ (66.97,68.54)\end{tabular}          & \begin{tabular}[c]{@{}c@{}}92.26\\ (91.52,92.85)\end{tabular}          \\
        RAD-DINO\cite{raddino}$^*$      & \begin{tabular}[c]{@{}c@{}}62.88\\ (61.38,64.26)\end{tabular}          & \begin{tabular}[c]{@{}c@{}}69.43\\ (68.50,70.32)\end{tabular}          & \begin{tabular}[c]{@{}c@{}}67.07\\ (66.33,67.77)\end{tabular}          & \begin{tabular}[c]{@{}c@{}}91.49\\ (90.65,92.21)\end{tabular}          \\
        BiomedGPT\cite{biomedgpt}$^\ddagger$ & \begin{tabular}[c]{@{}c@{}}64.71\\ (63.25,66.04)\end{tabular}          & \begin{tabular}[c]{@{}c@{}}71.92\\ (71.01,72.81)\end{tabular}          & \begin{tabular}[c]{@{}c@{}}68.63\\ (67.85,69.42)\end{tabular}          & \begin{tabular}[c]{@{}c@{}}92.86\\ (92.23,93.46)\end{tabular}          \\
        BiomedCLIP\cite{biomedclip}$^\ddagger$ & \begin{tabular}[c]{@{}c@{}}66.60\\ (65.19,67.86)\end{tabular}          & \begin{tabular}[c]{@{}c@{}}73.24\\ (72.36,74.15)\end{tabular}          & \begin{tabular}[c]{@{}c@{}}71.29\\ (70.56,72.01)\end{tabular}          & \begin{tabular}[c]{@{}c@{}}93.52\\ (92.95,94.00)\end{tabular}          \\ \midrule
        BioVFM                       & \textbf{\begin{tabular}[c]{@{}c@{}}69.92\\ (68.52,71.12)\end{tabular}} & \textbf{\begin{tabular}[c]{@{}c@{}}76.05\\ (75.09,76.92)\end{tabular}} & \textbf{\begin{tabular}[c]{@{}c@{}}73.43\\ (72.55,74.25)\end{tabular}} & \textbf{\begin{tabular}[c]{@{}c@{}}94.40\\ (93.90,94.84)\end{tabular}} \\ \bottomrule
    \end{tabular}
\end{table}

\subsection{Model scaling}
\begin{figure}[t]
    \centering
    \includegraphics[width=\linewidth]{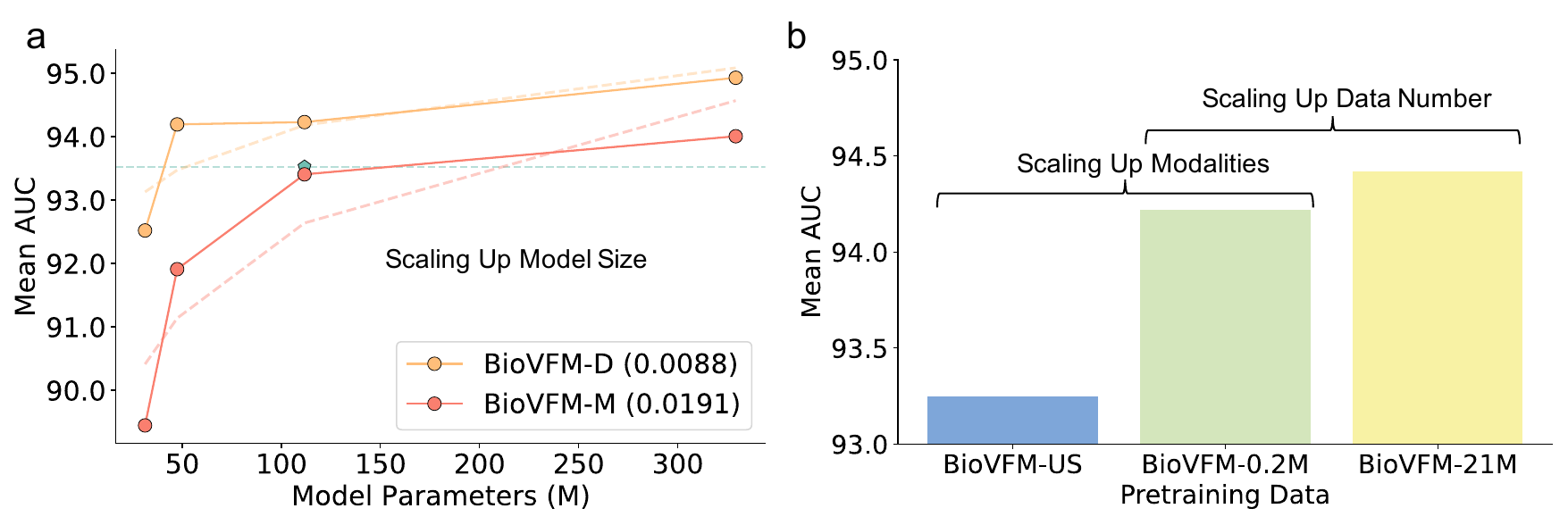}
    \caption{
        \textbf{Medical FMs scales with model size, data size, and number of modalities.}
        \textbf{a.} Model scaling of \method{}-D and \method{}-M. The dotted line indicates the fitted scaling curves and the value within brackets indicates the slope $a$. The turquoise star indicates the performance of BiomedCLIP. 
        \textbf{b.} Data and modality scaling with \method{}-D. The results are averaged over 12 medical benchmarks. 
    }
    \label{fig:model_data_scale_mean}
\end{figure}

\begin{figure}[t]
    \centering
    \includegraphics[width=\linewidth]{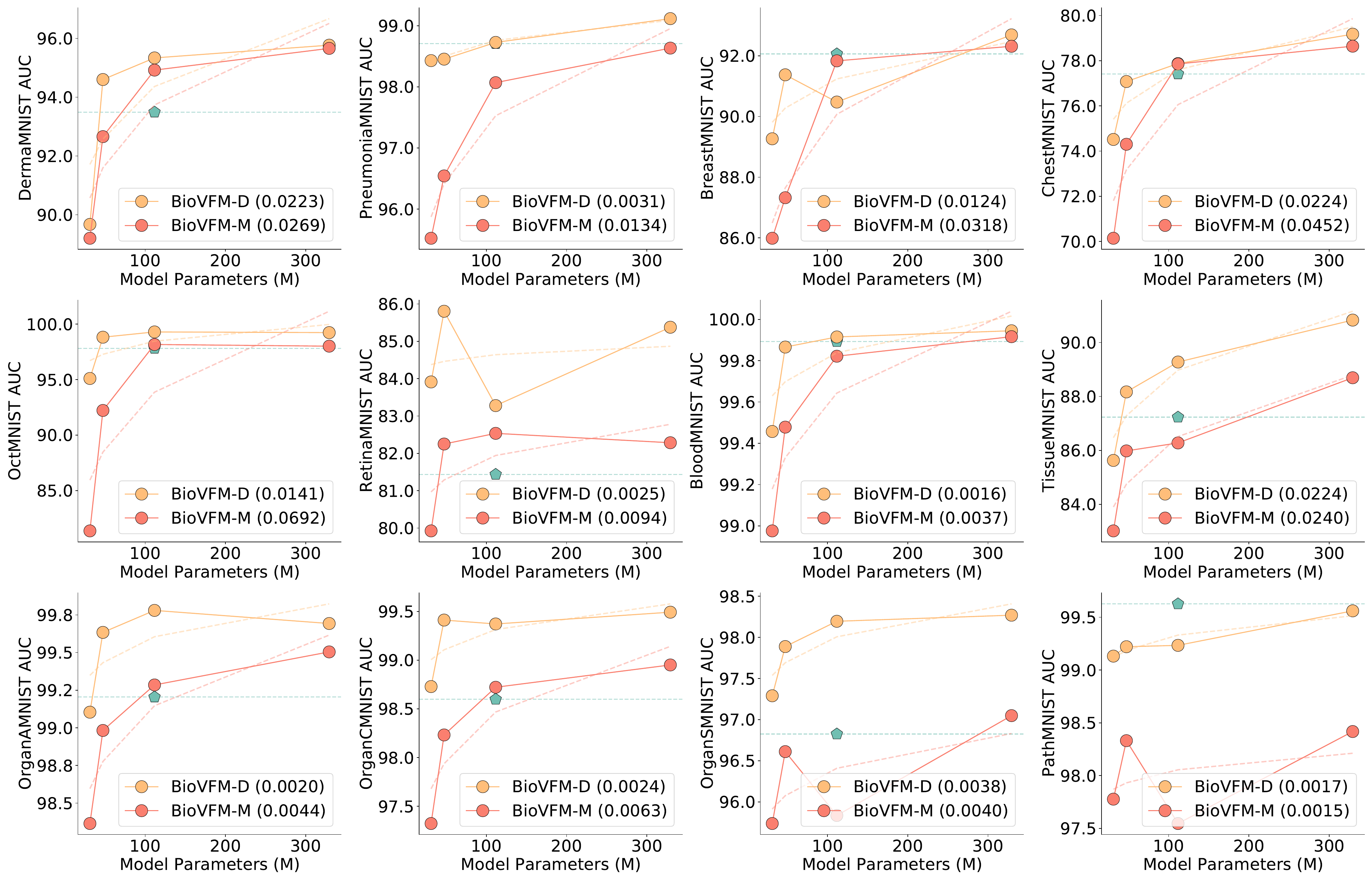}
    \caption{
        \textbf{Model scaling results across 12 medical benchmarks.} The dotted line indicates the fitted scaling curves and the value within brackets indicates the slope $a$. The turquoise star indicates the performance of BiomedCLIP.
    }
    \label{fig:model_data_scale}
\end{figure}

\begin{figure}[t]
    \centering
    \includegraphics[width=\linewidth]{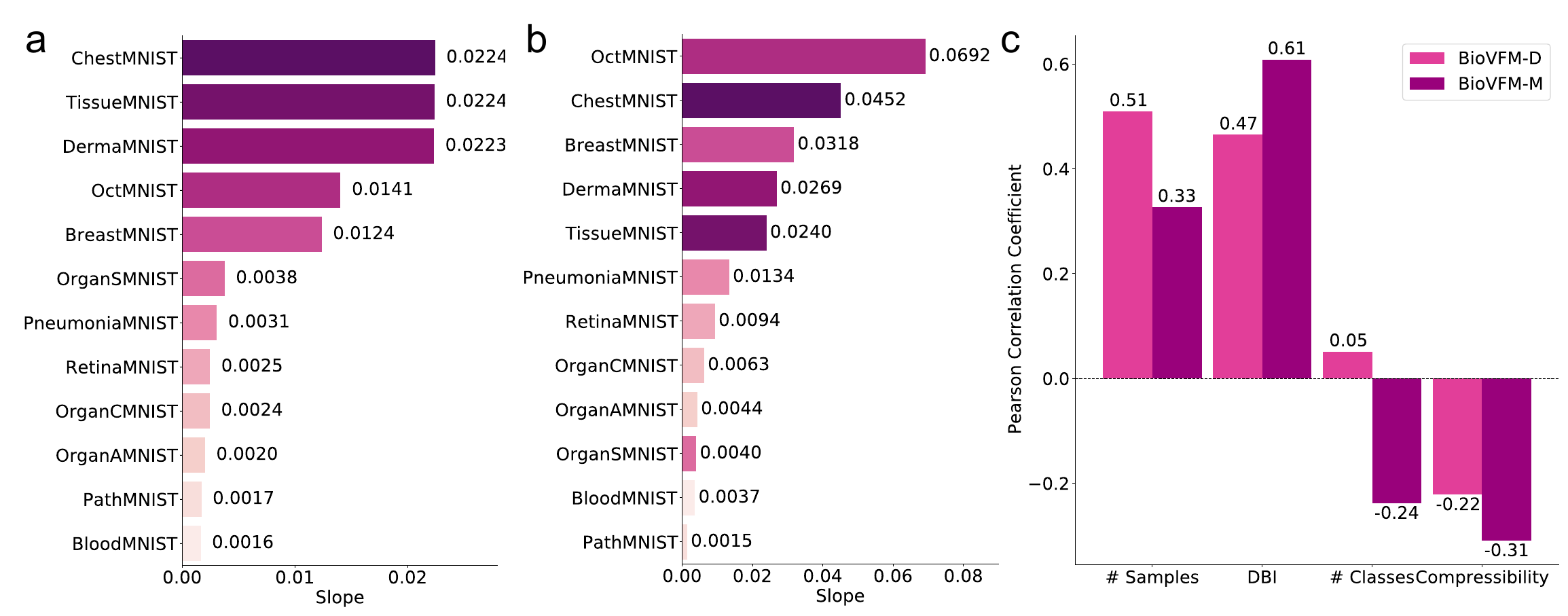}
    \caption{
        \textbf{Model scaling slopes and correlation to benchmark properties.}
        \textbf{a.} The model scaling slopes of \method{}-D with 12 medical benchmarks.
        \textbf{b.} The model scaling slopes of \method{}-M with 12 medical benchmarks. The colors are consistent for each task.
        \textbf{c.} The Pearson correlation coefficient between model scaling slopes and benchmark metrics.    
    }
    \label{fig:model_data_scale_slopes}
\end{figure}

As shown in Fig. \ref{fig:model_data_scale_mean}a and Fig. \ref{fig:model_data_scale}, when scaling up the model size from 5 to 303 million, both \method{}-D and \method{}-M demonstrate improved mean AUC as the scaling law predicted. The average performance of \method{}-D was improved by 2.1\% and \method{}-M was improved by 2.3\% through scaling up model size. 
However, scaling up model size does not guarantee consistent improvements across all medical tasks. As shown in Fig. \ref{fig:model_data_scale_slopes}, the scaling slope across different tasks and imaging modalities shows considerable variation, with differences exceeding 40-fold. For example, increasing model size from 5.5 to 303.3 million improves AUC by 4.65\% on ChestMNIST, whereas performance plateaus on RetinaMNIST when model sizes beyond 21.7 million parameters.

Interestingly, we observe a similar trend in the relative scaling slopes across different evaluation benchmarks. Specifically, for each benchmark, the computed scaling slopes may vary, but the relative order of the slopes remains stable:  the top 5 benchmarks with higher slopes consistently exhibit greater scaling benefits compared to those with lower slopes, regardless of the pretraining method used. The gap between high and low slope groups ranges from 3 to 14 folds, highlighting significant variability in scaling efficiency. However, estimating scaling benefits based on qualitative factors, such as imaging modality, anatomy, or view orientation, remains challenging.  For instance, although ChestMNIST and PneumoniaMNIST share the same imaging modality and anatomical region, they show different scaling slopes. OrganAMNIST, OrganCMNIST, and OrganSMNIST are derived from the same CT data but in different views, they demonstrate similar scaling slopes across all three tasks. These findings suggest that more nuanced factors, such as dataset size, task complexity, and intra-class variability, are likely to influence the scalability of vision foundation models.

\subsection{Data and modality scaling}
As shown in Fig. \ref{fig:model_data_scale_mean}b, scaling up image modalities leads to a noticeable improvement of 0.97\% in the mean AUC across 12 medical diagnosis tasks, demonstrating the importance of data diversity for enhancing generalizability. Surprisingly, we found that improving performance by scaling imaging modalities does not necessarily degrade the performance of specific tasks. The model trained on \method{}-0.2M outperforms the model trained on \method{}-US on the ultrasound benchmark (92.75\% vs. 91.87\% in AUC). Furthermore, simply scaling up the data size can yield diminishing returns, with only a 0.2\% improvement in mean AUC despite a 100-fold increase in data size. These findings suggest that data diversity, rather than just data size, plays a more significant role in developing scalable and generalizable medical foundation models.

\subsection{Scalability vary across pretraining algorithms}
We compare the self-supervised pretraining algorithms by two dimensions: the scaling efficiency (quantified by scaling slope) and the performance on medical benchmarks. As shown in Fig. \ref{fig:model_data_scale_mean}a, \method{}-M exhibits a steeper scaling slope (0.0191) compared to \method{}-D (0.0088), suggesting it benefits more from increased model capacity. However, as shown in Fig. \ref{fig:model_data_scale}, \method{}-D outperforms \method{}-M in almost all the benchmarks. The average gap between \method{}-D and \method{}-M across all benchmarks and model sizes is 1.7\%. Notably, \method{}-D outperforms BiomedCLIP on 7 out of 12 tasks even when using 3.5× fewer parameters. This divergence underscores the critical role of pertaining objective design in determining both scalability and downstream task performance when developing medical foundation models. 

\subsection{Correlations between benchmark properties and scaling slopes}\label{sec:scaling_corr}
To investigate the underlying impact factors related to scaling efficiency, we take model scaling as an example and perform a Pearson correlation coefficient analysis between quantitative metrics and scaling slopes. As shown in Figure \ref{fig:model_data_scale}c, the number of samples and DBI demonstrate a stronger correlation with model scaling slopes. A higher DBI indicates that the data has less distinct clusters, which increases the classification difficulty. This complexity necessitates larger models to capture complex decision boundaries, thus demonstrating better scaling efficiency. Additionally, the number of samples is positively correlated with model scaling slopes, as larger models can remember more information from more data to improve performance. Another meaningful indicator is g-zip compressibility, which measures the ratio of the compressed data size to the original size in bytes. Images with more redundancy will exhibit lower g-zip compressibility scores. Interestingly, we observed a negative correlation between g-zip compressibility and model scaling slopes, suggesting that larger models are more effective in handling image redundancy. In summary, larger models are advantageous for complex tasks, such as intricate decision boundaries and high data redundancy. However, simpler tasks should balance the diminishing returns of scaling with the associated computational overhead.

\section{Discussion}
This paper presents a systematic empirical study on the scalability of self-supervised learning for medical vision foundation models. We develop a large-scale medical image dataset to support this study. We investigate the scaling behavior over model sizes, data sizes, image modalities, and pretraining algorithms. Our results reveal that while scaling up did demonstrate benefits for medical foundation models, several important factors including pretraining algorithm, data diversity, or task complexity contribute to the varied scaling behaviors. We believe that this study paves the way for future medical AI systems that can adapt to a wide range of clinical tasks and settings. Finally, we propose \method{}, a medical vision foundation model at scale, demonstrating superior performance across 12 medical benchmarks. We will open the dataset, model, and algorithms of this study at \url{https://github.com/JiarunLiu/BioVFM}.

\begin{credits}
\subsubsection{\ackname} 
This research was partly supported by the National Natural Science Foundation of China (No. 62222118, No. U22A2040), Shenzhen Science and Technology Program (No. JCYJ20220531100213029), Shenzhen Medical Research Fund (No. B2402047), the Major Key Project of PCL under Grant PCL2024A06, National Key R\&D Program of China (No. 2023YFA1011400), Key Laboratory for Magnetic Resonance and Multimodality Imaging of Guangdong Province (No. 2023B1212060052), and Youth lnnovation Promotion Association CAS. 
\end{credits}

%
%
%
\bibliographystyle{splncs04}
\bibliography{myref}

\end{document}